\documentclass[11pt]{article}
\usepackage{nodalida2025}
\usepackage{marvosym}
\usepackage{todonotes}
\usepackage{times}
\usepackage{url}
\usepackage{latexsym}
\usepackage{array}
\usepackage{booktabs, multirow}
\usepackage{hyperref}
\usepackage{hyperref}
\hypersetup{
    breaklinks=true,
    hidelinks,
    pdfencoding=auto
}

\aclfinalcopy 

\title{Localizing AI: Evaluating Open-Weight Language Models for Languages of Baltic States}

\author{
    Jurgita Kapočiūtė-Dzikienė$^{1,2}$, Toms Bergmanis$^{3,4}$, Mārcis Pinnis$^{3,4}$\\[1ex]
    $^1$ Tilde IT, Lithuania\\
    $^2$ Faculty of Informatics, Vytautas Magnus University, Lithuania \\
    $^3$ Tilde, Latvia \\
    $^4$ Faculty of Computing, University of Latvia \\
    [1ex]
    \texttt{\{name.surname\}@tilde.com}
}

\date{}
\begin{document}
\maketitle
\begin{abstract}
Although large language models (LLMs) have transformed our expectations of modern language technologies, concerns over data privacy often restrict the use of commercially available LLMs hosted outside of EU jurisdictions.
This limits their application in governmental, defence, and other data-sensitive sectors. In this work, we evaluate the extent to which locally deployable open-weight LLMs support lesser-spoken languages such as Lithuanian, Latvian, and Estonian.
We examine various size and precision variants of the top-performing multilingual open-weight models, Llama~3, Gemma~2, Phi, and NeMo, on machine translation, multiple-choice question answering, and free-form text generation.
The results indicate that while certain models like Gemma~2 perform close to the top commercially available models, many LLMs struggle with these languages. Most surprisingly, however, we find that these models, while showing close to state-of-the-art translation performance, are still prone to lexical hallucinations with errors in at least 1 in 20 words for all open-weight multilingual LLMs.
\end{abstract}

\section{Introduction}
\label{intro}
Since the fall of 2022, OpenAI and other big tech companies have transformed LLMs from an obscure technology little known outside the academic circles to a major household name. 
Key to this was the LLMs' ability to perform tasks specified in free-form instructions, making them excel as NLP tools\footnote{\href{https://artificialanalysis.ai/leaderboards/models}{https://artificialanalysis.ai/leaderboards/models}.}.

Furthermore, these models can learn during inference from relevant examples provided as inputs, making them adaptable to new requirements or even tasks. \newcite{moslem-etal-2023-adaptive} showed that in such a setting, GPT-3, provided with relevant translation examples, outperforms machine translation systems of major companies, including Google, DeepL, and ModernMT.

However, data privacy concerns often constrain the use of commercially available LLMs hosted outside EU jurisdiction, limiting their application in governmental, defence, and data-sensitive private sectors. Fine-tuning and operational deployment of adapted models can incur prohibitive costs in the case of commercially available LLMs, emphasizing the need for sovereign AI solutions--locally deployable alternatives that ensure security, control, and compliance. 
Recently, many powerful alternatives to the commercially available online LLMs have emerged \cite{jiang2023mistral7b,dubey2024llama3herdmodels,gemmateam2024gemma2improvingopen,Mesnard2024GemmaOM,abdin2024phi3technicalreporthighly}. 
Although many of these LLMs officially support only a handful of languages with a large speaker base, their training data often incorporate texts from many other languages. 
Therefore, in practice, these languages receive some degree of support. 
However, the extent to which these languages are supported, to the best of our knowledge, still needs to be evaluated.

In this work, we aim to answer the question of to what extent, if at all, several popular open-weight models support Lithuanian, Latvian, and Estonian.
All three languages have relatively small speaker bases and thus are unlikely to be focal points of major multilingual open-weight LLMs.
We examine variants of Meta's Llama~3,  Google's Gemma2, Mistral's NeMo, and Microsoft's Phi3 in their performance in multiple-choice question answering (MCQA) and machine translation (MT).
We also manually assess the text quality generated by these models by identifying the rate of incorrect words when answering open-ended questions.

We find that while some models like Gemma~2 nearly match the performance of top commercial models, many LLMs struggle with these languages. Surprisingly, even those models that approach state-of-the-art translation capabilities are still susceptible to lexical hallucinations.

\section{Experimental Setting}

We evaluate LLMs on multiple-choice question answering, machine translation, and text generation quality in open-ended question answering. 
While our experiments focus on the model performance for three languages--Lithuanian, Latvian, and Estonian--each with a speaker base under 3 million, we also include results for Czech and English for comparison purposes.
We have chosen to evaluate models that consistently appear on various leaderboards.
Specifically, we assess the \textbf{8.03B} and \textbf{70.6B} parameter versions of \textbf{Llama~3} and \textbf{Llama~3.1}, as well as the \textbf{3.21B} parameter version of \textbf{Llama~3.2} \cite{dubey2024llama3herdmodels} from Meta; the \textbf{9.24B} and \textbf{27.2B} parameter versions of \textbf{Gemma~2} \cite{gemmateam2024gemma2improvingopen,Mesnard2024GemmaOM} by Google; the \textbf{3.8B} and \textbf{14B} versions of \textbf{Phi~3} by Microsoft \cite{abdin2024phi3technicalreporthighly}; and the \textbf{12.2B} parameter \textbf{NeMo} by Mistral~AI \cite{jiang2023mistral7b}. 
To provide context for our experiments, we include online models by OpenAI such as \textbf{GPT-3.5 Turbo} and \textbf{GPT-4o }\cite{openai2024gpt4technicalreport} and \textbf{DeepL} machine translation systems. 
In experiments assessing the quality of Lithuanian text generation, we incorporate the Lithuanian language-specific fine-tuned versions of Llama~2 \cite{touvron2023llama2openfoundation} with \textbf{7B} and \textbf{13B} parameters, developed by Neurotechnology -- \textbf{Lt-Llama 2} \cite{nakvosas2024openllama2modellithuanian}.

We run LLMs on our local hardware using the default inference parameters of the Ollama platform\footnote{\href{https://ollama.com/}{https://ollama.com/}.}, which offers several levels of precision for model quantization: 4bit, 8bit, and full-precision -- 16bit.
By default, we use \textbf{4bit precision} in all our experiments, albeit at the cost of some performance degradation. We also evaluate the performance drop due to quantization by contrasting the results of quantized models with their full-precision counterparts.

\paragraph{Machine Translation}
For MT experiments, we use the FLORES-200 benchmark dataset \cite{flores,nllb2022}, which comprises parallel sentences in over 200 languages\footnote{\href{https://github.com/facebookresearch/flores/tree/main/flores200}{https://github.com/facebookresearch/flores/tree/main/flores200}.}. 
We use the \textit{devtest} subset from FLORES-200, which contains 1,012 sentences. 
We test LLMs in a zero-shot inference scenario. We use the following English prompt to request text translations from the specified source and to the specified target language:
\begin{center}
    ``\textit{$\{lang_a\}$: $\{sentence_{lang_a}\}$\\Translate the above $\{lang_a\}$ text into $\{lang_b$\}\\$\{lang_b$\}: }''
\end{center}
The translation and evaluation are performed at \textit{sentence}-level; the inference is conducted in a single run for each test sentence.
For automatic evaluation of MT quality, we use COMET\footnote{\href{https://huggingface.co/Unbabel/wmt22-comet-da}{https://huggingface.co/Unbabel/wmt22-comet-da}.} \cite{rei-etal-2020-comet,rei-etal-2022-comet} as it has been shown to have a higher correlation with human judgments than BLEU \cite{papineni2002bleu} and to be more suitable for unrelated system comparison \cite{kocmi2024navigating}.

\begin{table*}[ht!]
\small
\center
\begin{tabular}{cc|cc||lcccc|c|cc|cc}
\toprule
      & \textbf{DeepL} & \multicolumn{2}{c||}{\textbf{GPT}} & \multicolumn{2}{l}{\textbf{Llama: 3}}               & \multicolumn{2}{c}{\textbf{3.1}} & \multicolumn{1}{c|}{\textbf{3.2}} & \textbf{NeMo}  & \multicolumn{2}{c|}{\textbf{Gemma~2}} & \multicolumn{2}{c}{\textbf{Phi~3}} \\
      & & \textbf{3.5-T}       & \textbf{4o}   & \textbf{8B}   & \textbf{70B} & \textbf{8B}  & \textbf{70B} & \textbf{3B}   & \textbf{12B} & \textbf{9B}           &\textbf{27B}         & \textbf{3B}     & \textbf{14B}    \\\midrule
\textbf{EN-LT} &0.92& 0.88            & 0.91      & 0.62 & 0.83   & 0.61 & 0.84     & 0.46  & 0.73     & 0.86         & \textbf{0.89}        & 0.26        & 0.32      \\
\textbf{EN-LV} &0.92& 0.88            & 0.91      & 0.59 & 0.82   & 0.58 & 0.83     & 0.44  & 0.72     & 0.83         & \textbf{0.88}      & 0.25        & 0.27      \\
\textbf{EN-ET} &0.93& 0.92            & 0.92      & 0.65 & 0.86   & 0.63 & 0.87     & 0.48  & 0.75     & 0.84         &\textbf{0.89}        & 0.27        & 0.37      \\
\textbf{EN-CS} &0.93& 0.91            & 0.92      & 0.81 & 0.90   & 0.82 & 0.90     & 0.66  & 0.84     & 0.89         & \textbf{0.91}        & 0.25        & 0.51      \\
\textbf{LT-EN} &0.87& 0.86            & 0.88      & 0.77 & 0.81   & 0.77 & 0.82     & 0.75  & 0.82     & \textbf{0.87}        & \textbf{0.87}        & 0.32        & 0.34      \\
\textbf{LV-EN} &0.89& 0.87            & 0.89      & 0.77 & 0.83   & 0.78 & 0.82     & 0.76  & 0.84     & 0.87         & \textbf{0.88}        & 0.33        & 0.34      \\
\textbf{ET-EN} &0.90& 0.90            & 0.90      & 0.78 & 0.83   & 0.79 & 0.82     & 0.76  & 0.85     & 0.88         & \textbf{0.89}       & 0.33        & 0.34      \\
\textbf{CS-EN} &0.89& 0.89            & 0.89      & 0.86 & 0.88   & 0.86 & 0.87     & 0.86  & 0.87     & \textbf{0.89 }        & \textbf{0.89}        & 0.32        & 0.33      \\
 \midrule
 \textbf{Avg.} & 0.91& 0.89& 0.90& 0.73& 0.85& 0.73& 0.85& 0.65& 0.80& 0.87& \textbf{0.89}& 0.29& 0.35
 \\

\bottomrule  
\end{tabular}
\caption{\label{mt_results} Automatic MT quality evaluation results in COMET scores across models and translation directions. DeepL and OpenaAI GPT 3.5-Turbo and 4o are provided for reference. Top results by open-weight models for each translation direction are marked in \textbf{bold}.}
\end{table*}

\begin{table*}[ht!]
\center
\small
\begin{tabular}{ccc||lcccc|c|cc|cc}
\toprule
      & \multicolumn{2}{c||}{\textbf{GPT}} & \multicolumn{2}{l}{\textbf{Llama: 3}}               & \multicolumn{2}{c}{\textbf{3.1}} & \multicolumn{1}{c|}{\textbf{3.2}} & \textbf{NeMo}  & \multicolumn{2}{c|}{\textbf{Gemma~2}} & \multicolumn{2}{c}{\textbf{Phi~3}} \\
       & \textbf{3.5-T}       & \textbf{4o}   & \textbf{8B}   & \textbf{70B} & \textbf{8B}  & \textbf{70B} & \textbf{3B}   & \textbf{12B} & \textbf{9B}           &\textbf{27B}         & \textbf{14B}    \\\midrule

\textbf{LT} & 0.734 & 0.941 & 0.607 & 0.768 & 0.618 & 0.834 & 0.435 & 0.715 & 0.861 & \textbf{0.898} & 0.001  \\
\textbf{LV} & 0.756 & 0.950 & 0.571 & 0.710 & 0.581 & 0.783 & 0.410 & 0.689 & 0.869 & \textbf{0.914} & 0.002  \\
\textbf{EN} & 0.903 & 0.962 & 0.883 & 0.938 & 0.872 & \textbf{0.947} & 0.740 & 0.898 & 0.931 & 0.943 & 0.886 \\
\textbf{ET} & 0.773 & 0.928 & 0.576 & 0.770 & 0.560 & 0.821 & 0.397 & 0.686 & 0.859 & \textbf{0.893} & 0.003  \\
\textbf{CS }& 0.818 & 0.937 & 0.769 & 0.888 & 0.743 & 0.892 & 0.676 & 0.800 & 0.907 & \textbf{0.910} & 0.296  \\ \midrule
\textbf{Avg.} & 0.797 & 0.944 & 0.681 & 0.815 & 0.675 & 0.855 & 0.532 & 0.758 & 0.885 & \textbf{0.912} & 0.238 \\
\bottomrule
\end{tabular}
\caption{\label{q_a_results} Automatic MCQA evaluation results measuring accuracies across models and languages. OpenaAI GPT 3.5-Turbo and 4o are provided for reference. Top results by open-weight models for each language are marked in \textbf{bold.}}
\end{table*}

\begin{table}[ht!]
\small
\begin{tabular}{lccc|cc}
\toprule
              & \multicolumn{2}{l}{\textbf{Llama~3.1}} & \textbf{3.2} & \multicolumn{2}{l}{\textbf{Gemma~2}} \\
\textbf{}     & \textbf{8B}       & \textbf{70B}       & \textbf{3B}  & \textbf{9B}      & \textbf{27B}      \\ \midrule
\textbf{$\Delta$MT}   & 0.074             & 0.009                  & 0.015        & 0.006            & 0.001             \\
\textbf{$\Delta$MCQA} & 0.100             & 0.058              & 0.004        & 0.010            & 0.000         \\ \bottomrule   
\end{tabular}
\caption{Performance drop (difference between \textit{Avg.} scores across all languages) for several 4bit models compared to their respective full precision versions on the two tasks -- MT (COMET points)  and MCQA (accuracy).} \label{tab:quanti}
\end{table}
\paragraph{Multiple-Choice Question Answering}
For MCQA experiments, we employ the Belebele dataset, a benchmark in multiple-choice machine reading comprehension~\cite{bandarkar-etal-2024-belebele}. 
This dataset pairs each question with a short passage from the FLORES-200 dataset. 
Each question includes four multiple-choice answers, with one being the correct option. 
The dataset consists of 900 questions involving 488 distinct passages, each linked to one or two related questions.
We use LLMs in a zero-shot inference scenario. We use the following English prompt where ``\textit{$\{context\}$}'', ``\textit{$\{question\}$}'' and ``\textit{$\{answer_\#\}$}'' are in a specific language (Latvian, Estonian, etc.):
\begin{center}
``\textit{This is the context: '$\{context\}$'. This is the question: '$\{question\}$'. Here are the 4 candidate answers: '1) $\{answer_1\}$'; '2) $\{answer_2\}$'; '3) $\{answer_3\}$'; '4) $\{answer_4\}$'. Report only the correct answer's ID (1, 2, 3, 4) using the mandatory JSON format: $\{answer\_id:^{\prime\prime}\}$.}''
\end{center}
The prompt explicitly requests the ID (e.g. `1', `2', `3', or `4') of the correct answer formatted in JSON. Our evaluation metric is accuracy.

\paragraph{Text Quality in Free Form Question Answering}
\label{question_answering}
To assess LLMs' ability to generate answers that adhere to Lithuanian and Latvian conventions and grammatical norms, we prompt models to answer ten free-form questions such as ``\textit{When did the Soviet Union collapse, how many new countries appeared, and what are their names?}'' and ``\textit{Provide a definition of artificial intelligence.}'' We conduct human evaluation by two Lithuanian and Latvian native speakers and linguistics experts.
We require evaluators to count text errors, mark grammatically incorrect words or incorrect inflexions, mark invented words not existing in the language, and mark words within syntactically incorrect sentence structures (see Table~\ref{tab:text_quality}).

We also assess whether the provided answers are factually correct. However, the factual accuracy results lack statistical significance due to the small sample size and should be interpreted with caution. For instance, it happened that GPT-4 answered all ten questions correctly for the Latvian language, but this outcome reflects a preliminary observation rather than a deep investigation. The results, therefore, should be viewed as part of a pilot study and not as definitive findings.

\begin{table*}[ht!]
\small
\center
\begin{tabular}{lrc||cc|c|cc}
\toprule
                    & \multicolumn{1}{l}{} & \textbf{GPT}     & \multicolumn{2}{c|}{\textbf{Llama} \textbf{3.1}} & \textbf{Gemma~2 }& \multicolumn{2}{c}{\textbf{Lt-Llama 2}} \\
                     &  & \textbf{4o}      & \textbf{8B}            & \textbf{70B }          & \textbf{27B}     & \textbf{7B}            & \textbf{13B}           \\
\midrule
 \multirow{3}{*}{\textbf{LT}}                   & \textbf{Words/Sentences}      & 320/38  & 724/68        & 300/24        & 1273/135 & 1132/69       & 1020/58       \\
                    & \textbf{Error Rate (\%)}                  & 3.44    & 12.98         & 7.67          & 4.08    & 1.15          & \textbf{0.98 }         \\
                    & \textbf{Answer acc.}          & 0.9     & 0.2           & 0.9           & 0.9     & 0.5           & 0.4           \\ \midrule
\multirow{3}{*}{\textbf{LV}} & \textbf{Words/Sentences}      & 1249/91 & 362/27        & 619/39        & 1171/97 & -             & -             \\
                    & \textbf{Error Rate (\%)}                  & \textbf{2.48}    & 18.51         & 11.31         & 5.98    & -             & -             \\
                    & \textbf{Answer acc.}          & 1       & 0.3           & 0.9           & 0.8     & -             & -            \\ \bottomrule
\end{tabular}
\caption{Human evaluation results for text generation quality in free form question answering.}\label{tab:text_quality}
\end{table*}

\section{Results and Discussion}

\paragraph{MT evaluation results} (see Table~\ref{mt_results}) demonstrate the Gemma~2 family as the most capable open-weight model family.
Gemma~2 27B emerges as the best locally deployable model, yielding COMET scores on par with OpenAI's GPT-3.5 Turbo and only marginally worse than GPT-4o.
Although specialised proprietary MT models like DeepL achieve the highest average score (0.91), freely available Gemma~2  models are not far off, with average COMET scores of 0.89 and 0.87 for the 27B and 9B versions, respectively.
In this context, the Llama family has little to offer, with the Llama~3.0 and 3.1 70B parameter models surpassing the much smaller Gemma~2 9B only in two out of eight translation directions. Smaller models, like Llama's 3B and 8B versions and Mistral's 12B NeMo, show equally meagre results given the high performance of Gemma~2 9B. Lastly, the results of Phi~3 prove that these models have very little support for multilingualism.
\paragraph{Quantisation impact on MT quality} analysis (see Table~\ref{tab:quanti}) reveals that while Llama models are negatively affected to some degree, the performance of full-precision models does not justify their use either. Increased inference time and memory requirements for the 70B model are too prohibitive unless several top-of-the-shelf enterprise-grade GPUs are available\footnote{\href{https://huggingface.co/blog/llama31}{https://huggingface.co/blog/llama31}.}. However, the full precision 8B parameter Llama~3.1 still does not reach the performance of the Gemma~2 9B 4bit version (0.80 vs 0.87). Gemma~2 family models, on the other hand, show a statistically insignificant drop in translation performance when the 4bit version is used, suggesting that their architecture is very robust to quantization.  

While the current MT results provide valuable insights into LLM capabilities, future work could benefit from more fine-grained error analysis using frameworks like MQM (Multidimensional Quality Metrics) and ESA (Error Span Annotation). These approaches allow detailed classification of errors-such as those related to accuracy, fluency, and terminology, and help quantify their impact on text usability. Incorporating these methods could provide deeper insights into model limitations and guide targeted improvements, particularly for smaller languages like Lithuanian, Latvian, and Estonian.

\paragraph{MCQA results} (see Table~\ref{q_a_results}) show that the Gemma~2 27B parameter model outperforms GPT-3.5 Turbo across all languages, coming second only to OpenAI's flagship model, GPT-4o. Notably, Gemma~2 27B achieves the highest accuracy among the open-weight models, outperforming Llama~3.1 and NeMo models for the Lithuanian, Latvian, Estonian, and Czech. The Phi models, however, perform poorly, particularly in non-English languages, and their results often fail to meet the required JSON output format, providing detailed responses instead of just answer IDs.
\vspace{-5pt}
\paragraph{Quantization impact on MCQA} analysis unveils a similar picture as the analysis above for MT: Llama models are more sensitive to quantization, while Gemma~2 are more robust. As a result, Gamma 2 models show little performance degradation when much more efficient 4bit models are used. It's worth noting, however, that the accuracy drops because quantization differs depending on the amount of data each language has. 
Less spoken languages like  Lithuanian, Latvian, and Estonian are affected more than English and Czech, for which overall results are better. For example, the Llama~3.1 70B model loses 0.06, 0.12, and 0.08 accuracy for Lithuanian, Latvian, and Estonian, respectively, but only 0.01 and 0.02 for English and Czech.

\vspace{-5pt}
\paragraph{Text Generation Quality} evaluation results (see Table~\ref{tab:text_quality}) show that most models produce more than one error per 100 words. The Lt-Llama~2 models, specifically fine-tuned for Lithuanian, are the exception, with an error rate of just 0.98\% and no invented words. Among multilingual models, OpenAI's GPT-4o achieves strong performance, with 0.94 grammatically incorrect or incorrectly inflected words per 100 for Lithuanian and 1.52 for Latvian, while generating a very small number of invented words (0.31 and 0.56 per 100 for Lithuanian and Latvian, respectively). In contrast, Llama~3.1 models show significant shortcomings, with the highest frequency of grammatical errors: 8.01 per 100 for Lithuanian and 11.88 for Latvian. Additionally, Llama~3.1 generates a substantial number of invented words: 4.28 per 100 for Lithuanian and 3.87 for Latvian. Gemma~2 models perform considerably better, with 2.36 grammatical errors per 100 for Lithuanian and 4.10 for Latvian, and fewer invented words: 0.39 per 100 for Lithuanian and 1.45 for Latvian. These findings highlight clear quality differences among models. While Llama~3.1's high error rates make it unsuitable for most commercial applications, Gemma~2 strikes a better balance, approaching GPT-4o's quality but still falling short. Notably, Lt-Llama 2 sets the strongest benchmark with near-perfect output, minimal grammatical errors, and no invented words. On average, users can expect at least one linguistic error in every 2-3 sentences from the best open-weight models like Gemma~2, or every sentence for models like Llama~3.1, unless further multilingual specialization becomes available.
\vspace{-5pt}

\paragraph{Lesser-spoken Languages} like Lithuanian, Latvian, and Estonian have less support in open-weight models compared to more populous languages such as Czech. These differences are more pronounced in smaller and lower-quality models, especially in tasks where models generate text in lesser-spoken languages (e.g., MT from English into Lithuanian). Comparatively good results for Czech suggest that these disparities are related to the amount of data each LLM has seen for each language during training, rather than factors such as the structural complexity of the language.

\section{Conclusions}

Our findings demonstrate that certain open-weight LLMs, such as the Gemma~2 family, achieve performance comparable to top-tier commercial products, such as general-purpose models like OpenAI's GPT-4o and specialized machine translation services like DeepL. This progress enables local, secure, and private solutions, supporting the development of sovereign AI for many language tasks in governmental, defence, and other data-sensitive private sectors. Nevertheless, unless specifically fine-tuned for languages like Lithuanian, most multilingual models are still surprisingly prone to lexical hallucinations, highlighting the need for 1)~high-quality language data for languages of the Baltic states and 2)~research on language-specialized LLMs.

\section*{Acknowledgments}

This research has been supported by the ICT Competence Centre (\href{www.itkc.lv}{www.itkc.lv}) within the project \textit{2.4 Daudzvalodīgs uzņēmuma informācijas semantiskās meklēšanas un atbilžu gatavošanas risinājums} (2.4 Multilingual Semantic Search and Question-Answering Solution for Enterprises) of EU Structural funds, ID no 5.1.1.2.i.0/1/22/A/CFLA/008.


~

\bibliographystyle{acl_natbib}
\bibliography{nodalida2025}

\end{document}